\def\year{2020}\relax
\documentclass[letterpaper]{article} 
\usepackage{aaai20}  
\usepackage{times}  
\usepackage{helvet} 
\usepackage{courier}  
\usepackage[hyphens]{url}  
\usepackage{graphicx} 
\usepackage{graphicx}  
\usepackage{eqnarray,amsmath}
\usepackage{booktabs}
\usepackage[small]{caption}
\title{CreditPrint: Credit Investigation via Geographic Footprints by Deep Learning}
\author{Xiao Han, Ruiqing Ding, Leye Wang, Hailiang Huang
}
 
\begin{document}

\maketitle

\begin{abstract}
Credit investigation is critical for financial services. Whereas, traditional methods are often restricted as the employed data hardly provide sufficient, timely and reliable information. With the prevalence of smart mobile devices, peoples' geographic footprints can be automatically and constantly collected nowadays, which provides an unprecedented opportunity for credit investigations. Inspired by the observation that locations are somehow related to peoples' credit level, this research aims to enhance credit investigation with users' geographic footprints. To this end, a two-stage credit investigation framework is designed, namely \textit{CreditPrint}. In the first stage, CreditPrint explores regions' credit characteristics and learns a credit-aware embedding for each region by considering both each region's individual characteristics and cross-region relationships with graph convolutional networks. In the second stage, a hierarchical attention-based credit assessment network is proposed to aggregate the credit indications from a user's multiple trajectories covering diverse regions. The results on real-life user mobility datasets show that CreditPrint can increase the credit investigation accuracy by up to 10\% compared to baseline methods.
\end{abstract}

\section{Introduction}


User credit investigation is one of the most fundamental services required by a variety of financial businesses such as credit card application, loan approval, and insurance pricing. In conventional credit industry, the credibility of credit card holders are often assessed by statistics of their credit card transaction records (e.g., FICO score\footnote{https://www.fico.com/en/products/fico-score}); credit investigations for people without any credit card mostly relies on users' self-reported information (e.g., income, occupation, and education) and public record information like tax liens and civil judgements~\cite{emekter2015evaluating}. Though existing credit investigations can provide credit scores and prevent severe financial frauds to some extent, the persistently high debt delinquency and default rates reveal the pitfalls. According to the latest loan debt statistics of U.S. for 2019\footnote{https://www.newyorkfed.org/newsevents/news/research/2019/20190514}, approximately 5.04\% of all credit card loans are 90-plus days past due, and the default rate (90 days or more delinquent) of student loan debt is reaching to 9.54\%.

Analytically, conventional credit investigations are primarily restricted to the deficiency of the employed data from several aspects:

(i) \textbf{Richness}. The information used for current credit investigations is still very limited and only reflects users' credits from partial aspects. Moreover, the information samplings hardly reach a full coverage of people. For instance, FICO score is not available for students without credit cards.

(ii) \textbf{Timeliness}. It is common sense that a user's credit status varies along time. For instance, the change of occupation such as losing job or getting promoted may lead to changes of credit status. However, it is quite hard for current credit investigations to timely update some important information if the user does not report the change. 

(iii) \textbf{Reliability}. For better credit investigation results, people may manipulate their reported information. For instance, one may request 
friends temporarily transfer an amount of money to her/him. Hence, the reliability of such self-reported information is also doubtful.


With the prevalence of ubiquitous computing equipments, such as smartphones, tablets, and smartwatches, more and more information can be continuously collected from our daily life. This offers a great opportunity to develop novel credit investigation techniques that may overcome the aforementioned limitations.
For instance, Sesame Credit\footnote{https://www.creditsesame.com/} leverages machine learning techniques to analyze users' credibility through their purchase and socializing behaviors in the Taobao website and app.
While pioneering credit investigation studies with ubiquitous computing devices mainly focus on users' behavioral data in virtual space, such as online borrowing, consuming, or entertaining~\cite{P2pLending}, 
this paper explores people's \textbf{geographic footprints} data from their physical space for credit investigation. Specifically, it investigates whether the geographic footprints of people (e.g., obtained from smartphones) can reveal characteristics about their credit, and how to predict users' credit levels with the footprint data.

\begin{figure}[t]
	\centering
	\includegraphics[width=0.48\textwidth]{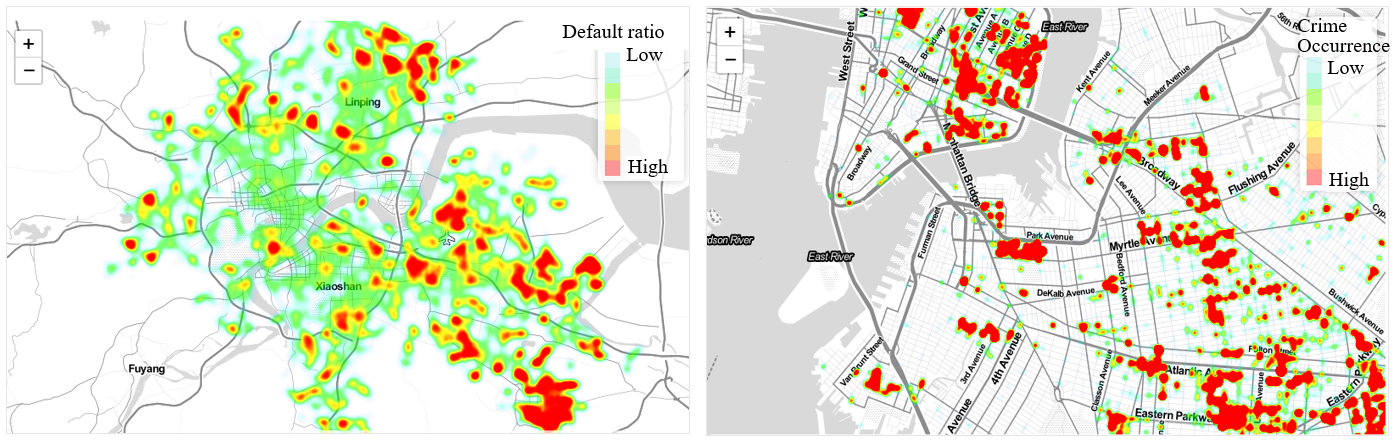}
	\caption{User credit geographic distribution (left, Hangzhou, China) and crime geographic distribution (right, New York, US).}
	\vspace{-1em}
	\label{fig:credit_distribution}
\end{figure}

Many clues drive us to conduct credit investigation through users' geographic footprints. Firstly, geographic footprints can be automatically and continuously sensed by up-to-date ubiquitous computing equipments with localization capacities (e.g., GPS). In other words, geographic footprints can be \textit{timely} collected with ensured \textit{reliability}. Moreover, an arising research line of work shows that users' daily trajectories or their frequently visited locations bear certain social and economic indications that can be used for many predictive applications, such as crime occurrence prediction~\cite{deepCrime} and real estate ranking~\cite{fu2016modeling}. More importantly, just as the correlation (e.g., the correlation between locations and crime occurrences) findings in the prior studies, our preliminary investigation reveals that locations and visitors' credit level are somehow related (Figure~\ref{fig:credit_distribution}). For instance, some locations are full of footprints from high-credit people while some other locations likely assemble low-credit visitors. These observations demonstrate the feasibility of credit investigation via geographic footprints and inspire us to uncover the credit characteristics embedded in locations which may increase data \textit{richness} for credit investigation. 



Nevertheless, leveraging users' daily geographic footprints for credit investigation is also non-trivial, facing the following \textbf{challenges}:

(i) While each location can hold various features (e.g., points-of-interests), few are directly related to the credits of the visited users, thus how to characterize the location from the credit investigation perspective is the first challenge.

(ii) A person may have multiple footprint trajectories that cover different sets of locations with diverse (or even conflicted) credit indications, thus how to merge the credit indications from different locations to generate an integrated credit investigation result is the second challenge.


To address these challenges, this paper makes the following contributions:

(i) To the best of our knowledge, this paper is one of the pioneering research works studying how users' geographic footprints can reveal their credit trustworthiness. The exploration of geographic footprints facilitates credit investigation with better richness, timeliness, and reliability.

(ii) This paper proposes a novel two-stage footprint-based credit investigation framework, namely \textit{CreditPrint}. In the first stage, a \textit{credit-aware region embedding network} is designed to learn credit-related representation for each geographic region considering various relationships between regions. In the second stage, a \textit{hierarchical trajectory-based credit assessment network} is proposed to fuse a user's multiple footprint trajectories covering diverse regions into an integrated representation for final user credit prediction.

(iii) With a real-life user mobility dataset with credit scores including more than 5,000 users, we verify that our proposed CreditPrint framework can significantly improve the credit prediction accuracy by more than 10\% compared to baseline methods.


\section{Related Work}

In literature, traditional credit investigation works usually leverage users' reported information, including income, occupation, education, etc., to predict users' credit scores \cite{zhao2017p2p}. Various learning techniques have been used for this purpose, including logistic regression \cite{emekter2015evaluating}, random forest \cite{malekipirbazari2015risk}, and neural network \cite{byanjankar2015predicting}. However, as illustrated in the introduction, these types of information may suffer from pitfalls in richness, timeliness, and reliability. Recently, researchers started to consider more users' data collected from pervasive computing devices. One work similar to ours is using users' mobile phone data to help credit assessment \cite{ma2018new}. Ma et al. (2018) manually extract some statistics about users' mobility patterns (e.g., mean and variance of the visited location number) as users' mobility features for credit investigation; the limitation is that the specific characteristics of each visited location is ignored. In comparison, we propose a region embedding network to extract credit-related representation for each location; also, our trajectory-based credit assessment network can learn hidden features useful for credit investigation beyond manually crafted ones.

With the popularity of ubiquitous and mobile devices, much research has been devoted to mining users' geographic footprints for various applications \cite{zheng2015trajectory}, such as location-based services \cite{yang2014modeling}, transportation management \cite{wang2019ridesharing,chen2018tripimputor}, and emergency response \cite{chen2018radar}. Particularly, our work can be seen as a novel footprint data classification task. Most classification tasks are relevant to explicit moving patterns that can be easily observed by ordinary people, e.g., whether a car turns round or not \cite{chen2018radar}; in comparison, our research aims to extract footprints' implicit patterns related to user credits, which is rather challenging.

The key techniques in CreditPrint are graph convolutional network \cite{defferrard2016convolutional} and attention \cite{vaswani2017attention} mechanisms. Literature has shown that graph convolutional network is powerful to encode spatial relationships \cite{geng2019spatiotemporal,chai2018bike}, and attention mechanisms can effectively extract temporal characteristics \cite{liang2018geoman}. Inspired by these works, we aim to combine graph convolutional network and attention mechanisms to extract implicit credit-related spatiotemporal factors from users' footprint data.

\section{Problem Formulation}
We begin with describing some preliminaries and formulating the user footprint-based credit investigation problem.

\textsc{Definition 1.} \textbf{\textit{Region.}} A spatial area $\mathcal{A}$ (e.g., a city) is partitioned into $m$ grids of equal-size (e.g., $1km\times1km$). A grid is a region, denoted by $r$. The set of regions belonging to a spatial area $\mathcal{A}$ is denoted by $\mathcal{R}_\mathcal{A}=\{r_1,r_2,..., r_b\}$.

\textsc{Definition 2:} \textbf{\textit{Trajectory.}} A trajectory $t$ is a user $u$'s temporal sequence of footprints in terms of regions for a predefined time period (e.g., one day). Then a user $u$'s trajectory at time period $k$ can be denoted as $t^{(k)}_u =\{r^{(k)}_1, r^{(k)}_{2}, ... r^{(k)}_{m}\}$, where $r^{(k)}_i$ indicates the visit region at timestamp of $i$ during time period $k$.

\textbf{Footprint-based credit investigation problem.} Suppose there are a set of users $\mathcal{U}$ and their credit levels $\mathrm{Y}=\{y_u|u\in \mathcal{U}\}$, where $y_u=1$ indicates that user $u$ is of low credit and $y_u=0$ presents that $u$ is of high credit. Each user $u$ holds a set of trajectory records $\mathcal{T}_u = \{t^{(1)}_u, t^{(2)}_u, ..., t^{(n)}_u\}$ for $n$ time periods. Given a new user $u'$ and his trajectory records $\mathcal{T}_{u'}$, our main task is to predict the credit level $y_{u'}$ of $u'$. More specifically, we aim to learn a prediction function $\hat{y}_{u'} = \mathcal{F}(u'|T, Y, \Theta)$, where $\hat{y}_{u'}$ denotes the predicted probability that user $u'$ is of low credit and will default in the future, $\mathcal{T} = \{\mathcal{T}_u|u\in \mathcal{U}\}$ collects all users' historical trajectories, and $\Theta$ is a set of model parameters of function~$\mathcal{F}$. 


\section{CreditPrint}
In this section, we propose a CreditPrint framework to address the footprint-based credit investigation problem. We first overview the proposed approach and then illustrate each component of the framework.

\subsection{Framework Overview}

\begin{figure}[t]
	\centering
	\includegraphics[width=0.48\textwidth]{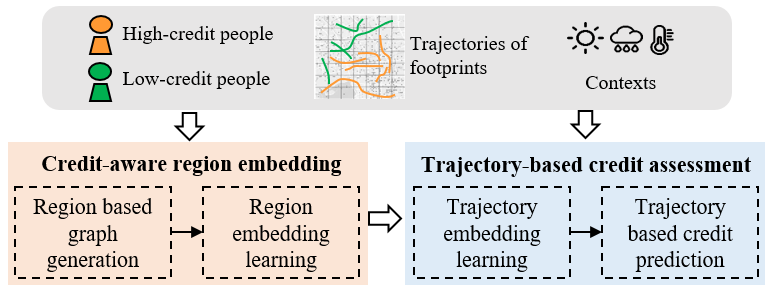}
	\caption{Framework of CreditPrint.}
	\vspace{-1em}
	\label{fig:framework}
\end{figure}

Intuitively, if a user frequently visits regions where many people of high credit travel, this user may be also of high credit, and vice-verse. On this basis, firstly, CreditPrint tends to generate an embedding for each region to characterize the credit worthiness of people who visit the region; and then CreditPrint represents a user's trajectories by sequences of credit-aware region embeddings and learns a user's credit score through his trajectories. In particular, CreditPrint encompasses two major components:

\textbf{(i) Credit-Aware Region Embedding.} As observed in Figure~\ref{fig:credit_distribution}, some regions are highly attractive to people with high credit score to visit, while some others often assemble footprints of people with low credit score. In other words, regions are of features that can indicate the visitors' credit score. Therefore, taking a collection of users' information including their historical trajectories and credit levels as input, the goal is to generate region embeddings for characterizing the regions in terms of credit worthiness. Straightforwardly, a region with more footprints of high-credit people and less footprints of low-credit people is more likely a high-credit region. In addition, regions may relate to each other in terms of credit worthiness by various ways. For instance, people of a certain credit level may regularly traverse two regions, which indicates that the two regions may share some credit characteristics. In this vein, we propose graph convolutional neural networks (GCN) to learn region embeddings by encoding both regions' individual characteristics and relationships in terms of credit worthiness.

\textbf{(ii) Trajectory-based Credit Assessment.} Once region embeddings are obtained, a trajectory can be easily represented by a sequence of region embeddings with credit characteristics. Gated Recurrent Units (GRUs) are naturally designed to model sequences and maintain sequential dependencies for a long-term. Then, we basically employ GRUs to learn trajectory embeddings by taking the sequences of credit-aware region embeddings as input. Ideally, if a trajectory of regions that traverses frequently by people with high/low credit score, then the trajectory embedding accordingly encodes the characteristics that indicate high/low credit state. In reality, a user often exhibits a number of trajectories. The collective trajectory embeddings of a user are expected to reveal whether the user is of high or low credit score. In particular, a user's trajectory may be affected by some context features (e.g., weather and traffic). Thus we attentively merge a user's collection of trajectory embeddings by taken into account the context features. Finally, a user embedding in terms of his trajectories and some manual-crafted trajectory features are concatenated to learn the user's credit state.

\subsection{Credit-Aware Region Embedding}

\begin{figure*}[t]
\centering
\includegraphics[width=0.95\textwidth]{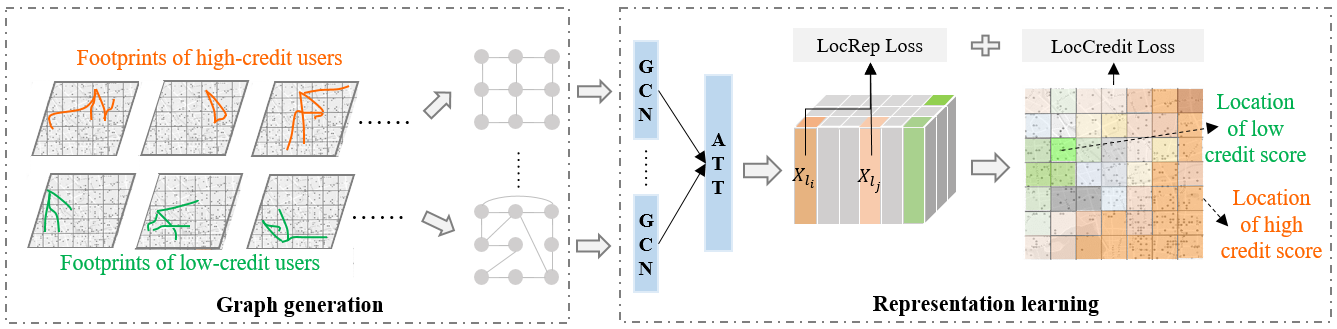}
\caption{REN: Credit-Aware Region Embedding by Graph Convolutional Neural Networks with Attention.}
\label{fig:locationEmbedding}
\end{figure*}
This section proposes a credit-aware Region Embedding Network (REN)  to generate region embeddings by considering both the regions' individual characteristics and the regional relationships in terms of credit worthiness. Specifically, we assign each region a credit score with respect to the synthetical credit levels of visitors in the region to describe its individual credit characteristics. In addition, concerning that GCN is a popular solution that can generate node embeddings by aggregating information from the nodes' neighbors, we leverage GCN to encode the complex relationships of credit worthiness between regions. As there are perhaps various relationships among various regions, we construct multiple graphs according to the various relationships and propose a credit-aware region embedding network based on GCN with attentions. Figure~\ref{fig:locationEmbedding} illustrates the framework of credit-aware region embedding.






\subsubsection{Region credit score.} We assign a credit score for each region to interpret the tendency of region assembling people of low credit. Therefore, the more users of low credit visited a region, the higher score it gains. Specifically, let respectively denote the number of visitors with low credit in a region $r$ by $|\mathcal{V}_r^-|$, and the overall number of visitors in the region $r$ by $|\mathcal{V}_r|$. We compute the region credit score $s_r$ by:
\begin{equation}
s_r = |\mathcal{V}_r^-|/|\mathcal{V}_r|
\end{equation}

\subsubsection{Graph Generation.} Graph generation is the primary task for graph convolutional operation. In our context, regions are the nodes in the graphs. Edges between any two regions are supposed to represent their relationships in terms of credit worthiness. And the edges are usually assigned with large weights if the connected two regions are of strong relationships. In particular, we propose three types of graphs to capture different types of relationships between regions.

\begin{itemize}
  \item \textbf{\textit{Region Distance Graph.}} According to Tobler's first law of geography: \textit{everything is related to everything else, but near things are more related than distant things}~\footnote{https://en.wikipedia.org/wiki/Tobler\%27s\_first\_law\_of\_geography}. It is also intuitive that people in a region are more likely to visit the nearby regions than the distant regions. In other words, a region maintains more similar footprints of people and present more similar credit characteristics to the surrounding regions than the distant ones. In this vein, we construct a distance graph by connecting adjacent regions.

  \item \textbf{\textit{Region Interaction Graph.}} In addition to visiting surrounding regions, people may also frequently traverse between certain distant regions due to various reasons. For instance, a user of high credit may travel from his home region to work region every work-day. Naturally, two regions are more interacted if users are more likely to traverse between them. Following this idea, we connect regions if they simultaneously appear to sufficient number of trajectories and build up an region interaction graph.

  \item \textbf{\textit{Region Correlation Graph.}} Despite some regions are not sequentially traveled by users, they are also perhaps strongly correlated to each other because of some similar regional attributes. For instance, many high credit users regularly go to some decent workplaces located in different regions in the morning of work day; and museums in different regions are  also often visited by high credit users during weekends. Therefore, we try to capture the regions' correlations by their similarity of footprint pattern. In particular, we construct a dynamic credit score vector to record the average ratio of low credit visitors to overall visitors by time (e.g., hour). Two regions are connected to build a region correlation graph if their dynamic credit score vector are correlated.
\end{itemize}

\subsubsection{Multiple Graph Convolution with Attentions.} In order to synthetically capture various types of relationships between regions, we learn to merge the graphs with attentions. Assume that we generate $M$ types of graphs $\mathcal{G} = \{G_1, G_2, ..., G_{_M}\}$ and $A_g$ represents the adjacent matrix of $G_g$. We add regions' self-connection on graph (i.e., $A_g \leftarrow A_g + I$) to ensure that the previous region embeddings can be involved in the update operations. In addition, we normalize the adjacent matrices to eliminate graph differences and hold stability of parameter learning. Then we have,
\begin{equation}
\tilde{A}_g = D_g^{-\frac{1}{2}} (A_g+I) D_g^{-\frac{1}{2}}
\end{equation}
where $D_g$ is a diagonal degree matrix with entries $D_g^{ii} = \sum_{j}A_g^{ij}$. Assume the weight of graph $g_q$ is represented by $W_g$. We try to normalize the graph weight and merge various graphs with attentions as:
\begin{align}
 a_g &= \text{softmax}(W_g) = \frac{\exp(W_g)}{\sum_{g=1}^{M}\exp(W_g)} \\
\tilde{A}_G &= \sum_{g=1}^{M} a_g \odot \tilde{A}_g
\end{align}
where $\odot  $ stands for element-wise product. After attentively merge the graphs, we conduct convolutional operations to update the hidden state by:
\begin{equation}
H^{(l+1)}_G = \sigma(\tilde{A}_G H^{(l)}_G W_G^l)
\end{equation}

\subsubsection{Loss Function of REN}
It is worth noting that there is a significant difference between our model and conventional graph convolution neural networks (GCN). In general, GCNs require a supervised training for obtaining the parameters by minimizing the errors between the actual and predicted labels with the following loss function:
\begin{equation}
\mathcal{L} = \sum_{r\in R} y_r \log \sigma(X_r^T \mathbf{\theta})+(1-y_r)\log(1-\sigma(X_r^T \mathbf{\theta}))
\end{equation}
where $y_r$ indicate the region credit label in our context. We set $y_r = 1$ if $s_r$ is larger than the median value of region credit scores, otherwise, $y_r=0$.

We notice that the regions with similar credit characteristics should also have similar embeddings while the embeddings of regions with diverse credit scores are supposed to be unlike. Therefore, we propose a region embedding similarity regularizer to supervise the learning. Specifically, we respectively use $R^+$ and $R^-$ to denote the high credit regions and low credit regions. We set a similarity threshold $\delta$ to identify similar region pairs. In brief, given a region $r$, its similar regions $r_s \in R_s$ satisfies two conditions: (1) $r$ and $r_s$ belong to the same region set, and (2) the credit score difference between $r$ and $r_s$  is smaller than $\delta$. We also sample the same number of dissimilar region pairs for each region, where $r$'s dissimilar regions $r_d \in R_d$ also satisfies two conditions: (1) $r$ and $r_s$ are from different region sets, and (2) the credit score difference between $r$ and $r_s$  is larger than $\delta$. Finally, we introduce the region similarity loss as:
\begin{equation}
\label{eq:region_sim_loss}
\mathcal{L}_r = \sum_{r\in R} \sum_{\substack{r_s\in R_s \\ r_d\in R_d}} \log \sigma(X_r,X_{r^s}) + (1-\log \sigma(X_r,X_{r^d}))
\end{equation}
where $\sigma(x_1,x_2) = \frac{1}{1+\exp(\langle x_1,x_2\rangle)}$ is a sigmoid function and $\langle x_1,x_2 \rangle$ is the inner product of $x_1$ and $x_2$. To train our model, the unified loss function can be written as:
\begin{equation}
\label{eq:ren_loss}
\mathcal{L}_{_\textit{REN}} = \mathcal{L} +\gamma_{_\textit{REN}} \mathcal{L}_r
\end{equation}

\subsection{Trajectory-based Credit Assessment}

\begin{figure}[t]
\centering
\includegraphics[width=0.48\textwidth]{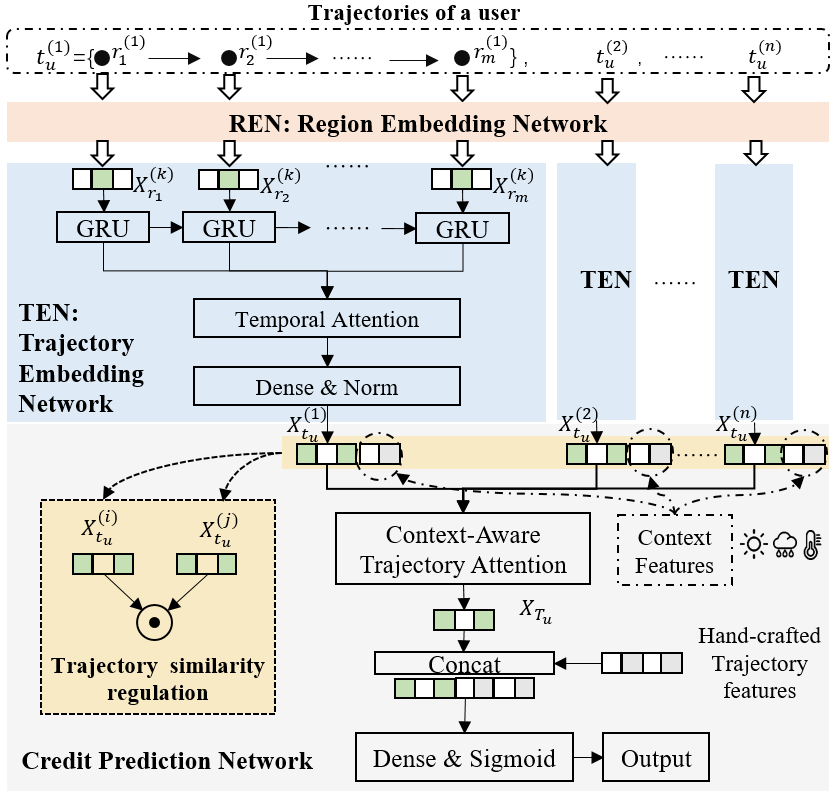} 
\caption{Trajectory-based Credit Assessment Networks.}
\label{fig:CreditPredict}
\end{figure}

In general, a user $u$ has a number of trajectories which contains multiple ordered regions. With the region embeddings that encode regions' credit characteristics, we represent $u$'s trajectories by sequences of region embeddings. Then, we propose a hierarchical \textit{Trajectory-based Credit Assessment Networks} (i.e., T-CAN), which firstly encodes each trajectory of $u$ by \textit{Trajectory Embedding Networks} (i.e., TEN) and then predicts $u$'s credit level with all of his trajectory embeddings by \textit{Credit Prediction Networks}. Figure~\ref{fig:CreditPredict} shows the framework of T-CAN.

\subsubsection{Trajectory Embedding Network}
Since a trajectory is represented as a sequence of region embeddings, TEN employs GRU, which is one of the most popular sequence model, to carry out trajectory embedding. A basic GRU the following update functions:
\begin{align}
  q_{r_i} = & \sigma(W^q X_{r_i} + U^q z_{r_{i-1}}) \\
  \gamma_{r_i} = & \sigma(W^{\gamma} X_{r_i} + U^{\gamma} z_{r_{i-1}}) \\
  h_{r_i} = & \text{tanh}(W^h X_{r_i} + U^h(z_{r_{i-1}}\odot \gamma_{r_i}))\\
  z_{r_i} =  & (1-q_{r_i})\odot h_{r_i} + q_{r_i}  \odot z_{r_{i-1}}
\end{align}
where $X_{r_i}$ is the $i$-th input region embedding of a trajectory, $z_{r_i}$ is the hidden state, $q_{r_i}$ is the update gate, $h_{r_i}$ is the candidate activation, $\sigma(\cdot)$ represents the sigmoid function, and $W^q$, $W^{\gamma}$, $W^h$, $U^q$, $U^{\gamma}$, $U^h$ are learnable parameters.

It is worth noting that the regions in a trajectory may present different importance when interpreting the trajectory's credit characteristics. For instance, the last region (i.e., destination) of a trajectory is often regarded more important than the region passed by in the middle, as the credit embedding of destination is more representative of the trajectory's credit characteristics. Therefore, we introduce temporal attention function to differentiate the inputs' weight as:
\begin{equation}
 a_{r_i} =  \frac{\exp(X_{r_i}W_t)}{\sum_{j=1}^{m} \exp(X_{r_j}W_t)}
\end{equation}
where $a_{r_i}$ is the attention score of $i$-the region of a trajectory, $z_{r_i}$ is the hidden state and $W_t$ is a trainable parameter. The training process is supervised by users' credit level. In this light, the region embeddings that are more representative to a user's credit level are supposed to be assigned with a higher score. The attention score is used to control the update of hidden state, which can be rewritten as follows:
\begin{equation}
z_{r_i} = (1-a_{r_i})z_{r_{i-1}}+ a_{r_i}h_{r_i}
\end{equation}
Taking the last hidden state $z_{r_m}$ of a trajectory $t$ as input, a dense network is used to output $t$'s embedding $X_t$.

\subsubsection{Credit Prediction Network}
Ideally, each of a user $u$'s trajectory embeddings is encoded of $u$'s credit characteristics and can be used to predict $u$'s credit level. Although $u$'s credit characteristics are relatively stable, $u$'s trajectories are often highly affected by surrounding context such as weather, wordday/weekend, or traffic, which may lead a great variance to $u$'s trajectory embeddings. Therefore, we take several measures to mitigate the variance due to context features, learn the stable credit characteristics and accurately predict $u$'s credit level.

Firstly, we represent a trajectory's surrounding context by a one-hot encoding feature vector and incorporate the context features to the trajectory's embedding. This incorporation could not only supervise trajectory embedding learning but also benefit for the credit level prediction. Secondly, to avoid prediction bias of a single trajectory, we aggregate $u$'s trajectory embeddings to synthetically predict $u$'s credit level. Thirdly, as a user's essential credit characteristics are often stable in periods of time, we regularize the similarity a user's two trajectories to further supervise the trajectory embedding process. We will elaborate the trajectory similarity regulation in the next subsection.

Let denote the context feature of trajectory $t^{(i)}_u$ by $C^{(i)}_{t_u}$. After concatenating $X^{(i)}_{t_u}$ and $C^{(i)}$, we aggregate a user' trajectory embeddings to obtain a user embedding $X_{T_u}$ by the following attention score:
\begin{align}
  a^{(i)}_{t_u} = & \frac{\exp(\text{Concat}(X^{(i)}_{t_u},C^{(i)}_{t_u}) W_u)}{\sum_{j=1}^{n} \exp(\text{Concat}(X^{(j)}_{t_u},C^{(j)}_{t_u}) W_u)}\\
  X_{T_u} =  & \sum_{i = 1}^n a^{(i)}_{t_u} X^{(i)}_{t_u}
\end{align}
where $n$ is the number of $u$'s trajectories and $W_u$ is a trainable parameter. Since a few manual-crafted trajectory features can significantly differentiate the high and low credit users, we also concatenate $u$'s manual-crafted trajectory features to his embedding $X_{T_u}$. A dense network with ReLU activation is then used to predict $u$'s credit label.

\subsubsection{Loss Function of T-CAN} To predict a user's credit level, we simultaneously train the trajectory embedding networks and the credit prediction networks. The primary goal of training is to minimize the user credit prediction error, i.e., logarithmic loss between the users' actual and predicted credit level:
\begin{equation}
\mathcal{L}_c = \sum_{u\in U} y_u \log \sigma(X_{T_u} \mathbf{\theta})+(1-y_u)\log(1-\sigma(X_{T_u} \mathbf{\theta}))
\end{equation}
In addition, users' trajectory embeddings are expected to encode his credit characteristics. As a user's credit characteristics are relatively stable, the embeddings of a user's different trajectories are preferably similar. Accordingly, we introduce a trajectory similarity regularizer to supervise the trajectory embedding process:
\begin{equation}
\label{eq:trj_sim_loss}
\mathcal{L}_{t_u} = \sum_{u\in\mathcal{U}} \sum_{\substack{i\neq j \\ t_u^{(i)}, t_u^{(j)}\in T_u }} \log \sigma(X^{(i)}_{t_u},X^{(j)}_{t_u})
\end{equation}
where $X^{(i)}_{t_u}$ and $X^{(j)}_{t_u}$ are embeddings of any two of $u$'s trajectories. Ideally, the trajectory similarity regularizer can direct the learner to extract a user $u$'s essential credit characteristics reflected by $u$'s various trajectories. Combining the user credit prediction error and trajectory similarity regularization, the complete loss function of T-CAN is written as:
\begin{equation}
\label{eq:tcan}
\mathcal{L}_{_{\textit{T-CAN}}} = \mathcal{L}_c +\gamma_{_{\textit{T-CAN}}} \mathcal{L}_{t_u}
\end{equation}

\section{Experiment}

In this section, we leverage real-life user mobility datasets to verify the effectiveness of CreditPrint. We first describe the experiment setup and then report the evaluation results.

\subsection{Experimental Setup}
\subsubsection{Data.} We use a user mobility dataset including more than 5000 users' trajectories and their credit label (whether they pay their mobile phone bill or not) for our experiment. The data is collected from Hangzhou, China by one mobile operator. The region is set to 1km*1km, and the data spans for one month. Figure~\ref{fig:credit_distribution} (left) gives an overview of our experiment area. We use 80\% of users as training, 10\% as validation, and 10\% as test.

\subsubsection{Benchmarks.} To compare with our method, we implement the following benchmark approaches. Assessing users' credit from footprints is still a very novel direction, and pioneering related work usually learns it from manually crafted features \cite{ma2018new}. Accordingly,  we design the following benchmark approaches:
\begin{itemize}
	\item \textit{Manual-LR}:  Using logistic regression \cite{emekter2015evaluating} with manual features for credit prediction.
	\item \textit{Manual-RF}: Using random forest \cite{malekipirbazari2015risk} with manual features for credit prediction.
	\item \textit{Manual-NN}: Using neural network \cite{byanjankar2015predicting} with manual features for credit prediction.
\end{itemize}
The manual features used in the experiment are summarized in Table~\ref{tbl:manual_features}. In addition, we  implement several variants of CreditPrint by removing some parts of the framework to verify that every proposed part can contribute to CreditPrint.

\begin{itemize}
	\item \textit{CreditPrint without REN}: This approach does not use the region embedding network (REN), but directly use the credit score (probability) of each region as their feature. The other parts are same as CreditPrint.
	
	\item \textit{CreditPrint without TEN}: This approach does not use trajectory embedding network (TEN), but directly use the average region embedding of a user's visited regions as the trajectory representation. The other parts are same as CreditPrint.
\end{itemize}

\begin{table}[t]
	\footnotesize
	\begin{tabular}{@{}ll@{}}
		\toprule
		\textbf{Name} & \textbf{Description} \\ \midrule
		\textit{num\_daily\_region} & number of visited regions per day \\ \midrule
		\textit{std\_daily\_region} & \begin{tabular}[c]{@{}l@{}}standard deviation of visited regions per\\ day\end{tabular} \\ \midrule
		\textit{region\_entropy} & \begin{tabular}[c]{@{}l@{}}entropy calculated by the region visiting\\ frequency\end{tabular} \\ \midrule
		\textit{turning\_radius} & \begin{tabular}[c]{@{}l@{}}mean distance of the visited regions to the\\ most-frequently-visited region\end{tabular} \\ \midrule
		\textit{weekday\_weekend\_diff} & \begin{tabular}[c]{@{}l@{}}difference between the number of visited\\ regions in weekday and weekend\end{tabular} \\ \midrule
		\textit{num\_record\_days} & \begin{tabular}[c]{@{}l@{}}number of days with at least one visited\\ region\end{tabular} \\ \bottomrule
	\end{tabular}
	\caption{Manual features used in experiments}
	\label{tbl:manual_features}
\end{table}

\subsubsection{Metrics.} We adopt the widely-used binary classification metric, AUC (Area Under Curve) to verify the effectiveness of user credit investigation.

\subsubsection{Network Structure.} Our region embedding network uses two graph convolutional layers each with 32 hidden units, and thus outputs a 32-dimension hidden vector as region representation. The trajectory embedding network generates a 128-dimension trajectory representation with one layer of GRU. We use such a design of the network structure because it performs well in the validation data set.


\begin{figure*}[t]
 \begin{minipage}{.2\textwidth}
  \scriptsize
  {\centering
   \begin{tabular}{@{}ll@{}}
    \toprule
    \textbf{Method} & \textbf{AUC} \\ \midrule
    \textit{Manual-LR} & 0.701 \\
    \textit{Manual-RF} & 0.695 \\
    \textit{Manual-NN} & 0.707 \\ \midrule
    \textit{CreditPrint without REN} & 0.732 \\
    \textit{CreditPrint without TEN} & 0.723 \\
    \textit{\textbf{CreditPrint (full model)}} & \textbf{0.784} \\ \bottomrule
   \end{tabular}
   \vspace{-.5em}
   \captionof{table}{Evaluation results}
   \label{tbl:evaluation}
  }
 \end{minipage}
 \begin{minipage}{.2\textwidth}
  \centering
  \includegraphics[width=1\linewidth]{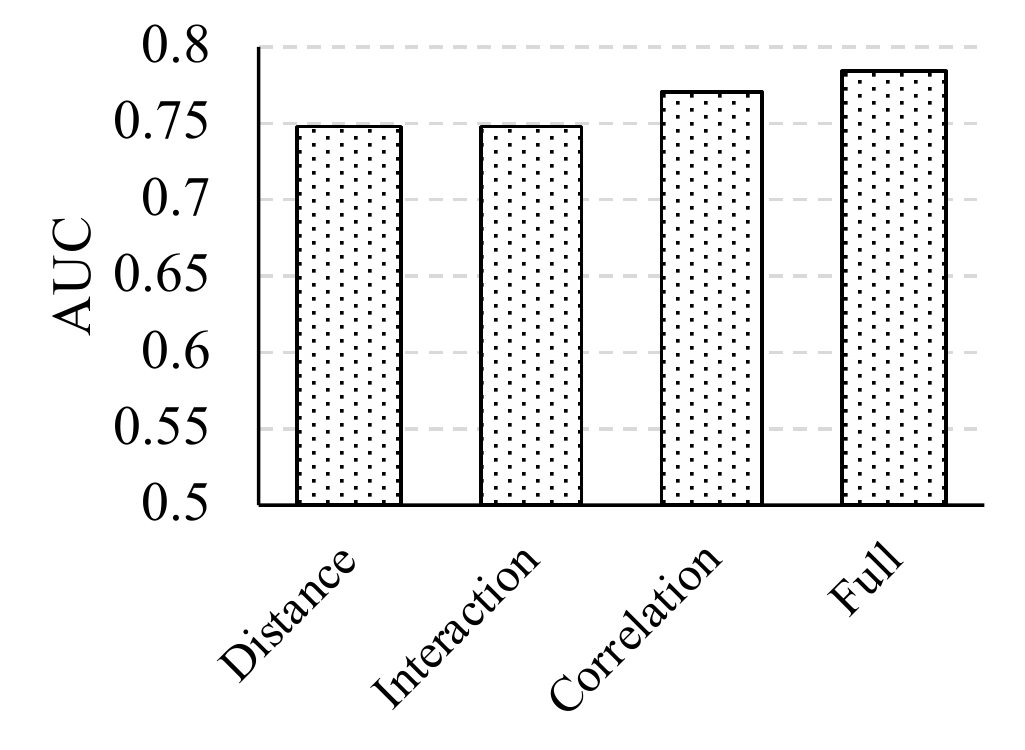}
  \caption{Region graphs}
  \label{fig:graph}
 \end{minipage}
 \begin{minipage}{.2\textwidth}
	\centering
	\vspace{+.8em}
	\includegraphics[width=.98\linewidth]{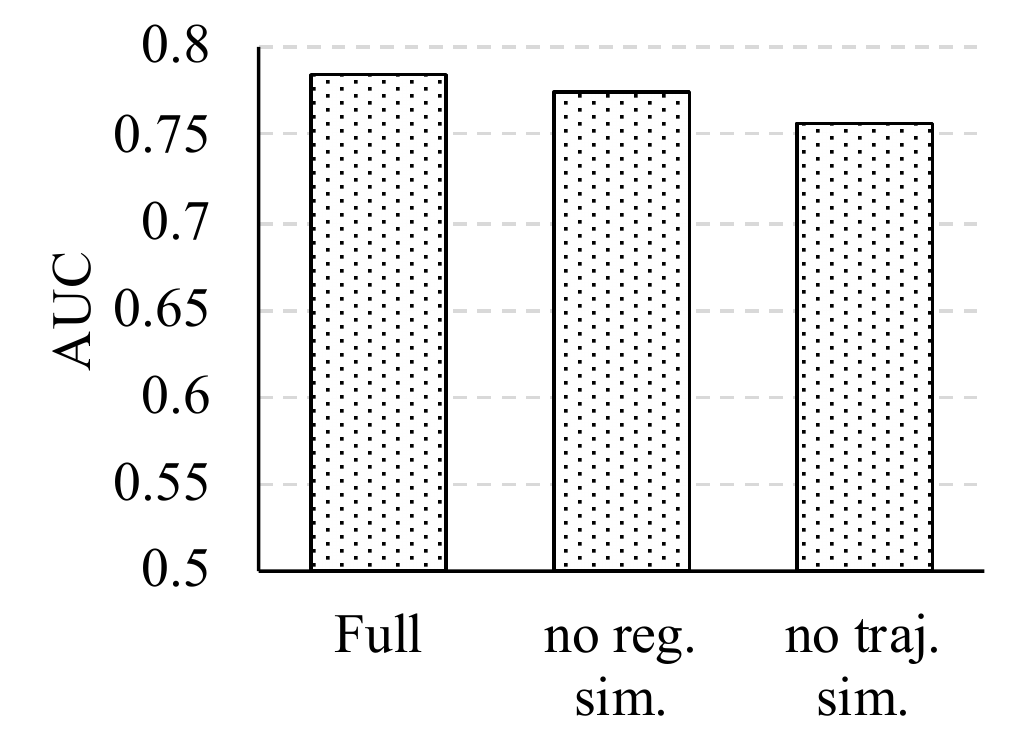}
	\caption{Region/Trajectory similarity regularizer.}
	\label{fig:sim_regularizer}
 \end{minipage}
 \begin{minipage}{.4\linewidth}
  \centering
  \includegraphics[width=0.49\linewidth]{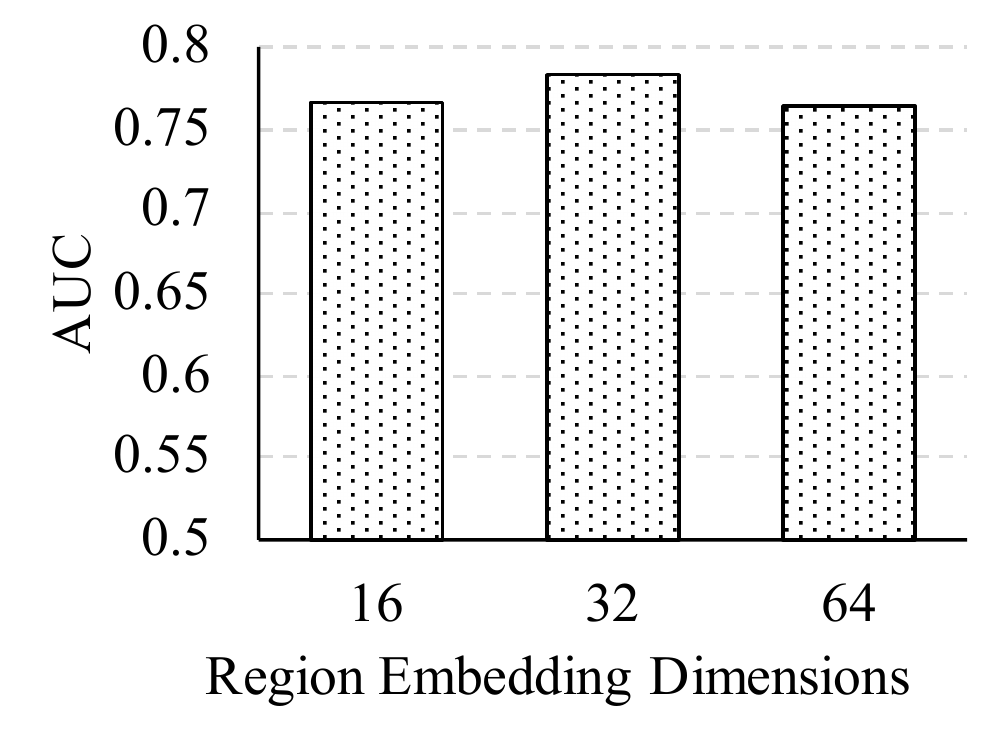}
  \includegraphics[width=0.49\linewidth]{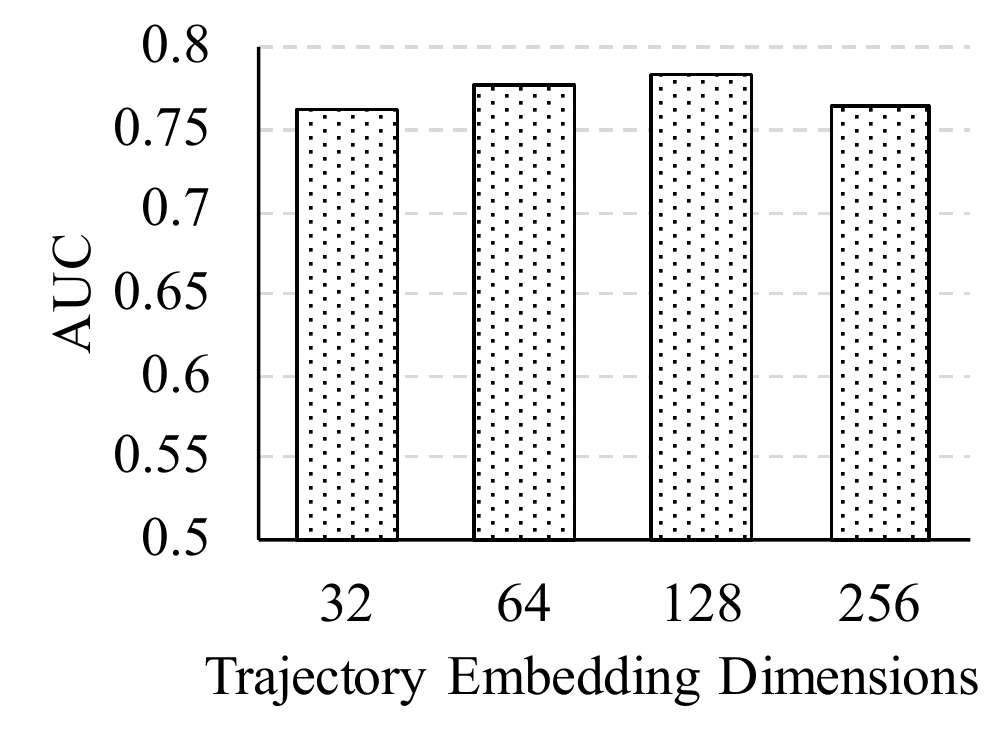}
  \caption{Region/Trajectory Embedding Dimensions.}
  \label{fig:region_emb_dim}
 \end{minipage}
\end{figure*}
\subsection{Results}

Table~\ref{tbl:evaluation} shows the evaluation results of CreditPrint and benchmark approaches. First, we can see all of the CreditPrint variants perform significantly better than manual-feature-based benchmark approaches. This indicates that considering the characteristics of a user's visited locations can actually give more credit-related information besides the mobility features manually extracted in literature. Particularly, our full CreditPrint model can increase AUC by more than 10\% compared to manual feature based benchmarks.

Second, we observe that our full CreditPrint model can also outperform other CreditPrint variants, verifying the effectiveness of each component designed in CreditPrint. Particularly, compared to CreditPrint without REN/TEN, our full model can increase AUC by around 7.1/8.4\%. This result highlights that both region embedding and trajectory embedding of CreditPrint are important for final credit assessment. Next, we verify more detailed design considerations in CreditPrint.

\subsubsection{Region Graph.} To verify the effectiveness of combining multiple region graphs (distance, interaction, and correlation graphs), we use only one graph to learn the region embedding for final credit assessment and compare to our full model with three graphs. Figure~\ref{fig:graph} shows the results and reveals that, among three graphs, the correlation graph can contribute most to the credit investigation. It also demonstrates that combining the three graphs together can further improve the prediction performance in terms of AUC.

\subsubsection{Region and Trajectory Similarity Regularizer.} In CreditPrint, we have introduced the region similarity regularizer (Eq.~\ref{eq:region_sim_loss}) and trajectory similarity regularizer (Eq.~\ref{eq:trj_sim_loss}) to assist the region and trajectory embedding learning. To verify their effects, we perform the experiment by removing the region similarity regularizer (set $\gamma_{_\textit{REN}}=0$ in Eq.~\ref{eq:ren_loss}) or the trajectory similarity regularizer (set $\gamma_{_\textit{T-CAN}}=0$ in Eq.~\ref{eq:tcan}), and then check the credit prediction results (Figure~\ref{fig:sim_regularizer}). Particularly, by removing the region similarity regularizer, AUC is reduced from 0.784 to 0.775; by removing the trajectory similarity regularizer, AUC is reduced from 0.784 to 0.756. These results verify that both region and trajectory similarity regularizers can help learn better representation for a user to reflect her/his credit level.

\subsubsection{Dimension of Region and Trajectory Embedding.} We briefly show how the results change according to the change of the dimensions of region and trajectory embedding in Figure~\ref{fig:region_emb_dim}. In general, we can see that the AUC does not change a lot by varying the dimensions of region embedding or trajectory embedding. We choose the region embedding dimension as 32, and the trajectory embedding dimension as 128, as these settings perform better than the others.

\section{Discussion and Conclusion}

This paper proposes a new credit investigation framework leveraging users' geographic footprint data, called \textit{CreditPrint}. The novelty of CreditPrint lies in two aspects. First, a credit-aware region embedding network is proposed to encode regions' characteristics and cross-region relationships to generate a credit-related region embeddings. Second, a hierarchical credit assessment network is constructed to learn a user's trajectory embedding from region embeddings, and further merge a user's multiple trajectory embeddings into an integrated user footprint feature vector with attentions for credit assessment. Evaluation results on real user mobility dataset verify the effectiveness of CreditPrint compared to the traditional manual crafted feature based methods.

It is worth noting that, while this paper focuses on using users' geographic footprints for credit investigation, the CreditPrint framework can be easily combined with other features that commonly used for credit assessment, such as income and education. Specifically, once we have such users' information, we can simply put it together with the hand-crafted footprint features and concatenate them to the users' footprint embeddings, as shown in Figure~\ref{fig:CreditPredict}.

Our research can also have implications for other research works depending on geographic footprint data mining. For example, persons' moving patterns have been revealed to be correlated to their mental health \cite{huang2016assessing}. Similar to CreditPrint, we may develop a framework to first learn regions' mental health-related embeddings and then extract users' footprint features by merging these region embeddings. This may enrich features for mental health prediction in \cite{huang2016assessing}.

In the future, we plan to extend CreditPrint in several aspects. First, we will try to find more scenarios, e.g., peer-to-peer lending, to evaluate whether users' geographic footprints can also indicate their credit well. Second, we will explore how to transfer the credit investigation model learned by CreditPrint from one area to another area, so as enhance its generalizability and applicability.

\bibliographystyle{aaai}
\bibliography{creditprint}

\end{document}